\documentclass[journal]{IEEEtran}

\usepackage{cite}
\usepackage{graphicx}
\usepackage{subfig}
\usepackage{amsmath}
\usepackage{multirow}
\usepackage{caption}
\usepackage{url}
\usepackage[colorlinks=false]{hyperref}
\usepackage{array}
\newcolumntype{*}{>{\global\let\currentrowstyle\relax}}
\newcolumntype{^}{>{\currentrowstyle}}
\newcommand{\rowstyle}[1]{\gdef\currentrowstyle{#1}%
	#1\ignorespaces
}

\begin{document}
%
\title{Direct Multitype Cardiac Indices Estimation via Joint Representation and Regression Learning}
%
%
%

\author{Wufeng~Xue,~Ali~Islam,~Mousumi~Bhaduri,~and Shuo~Li*%
\thanks{Copyright (c) 2017 IEEE. Personal use of this material is permitted. However, permission to use this material for any other purposes must be obtained from the IEEE by sending a request to pubs-permissions@ieee.org.}%
\thanks{W. Xue, A. Islam, M. Bhaduri and S. Li are with the Department of Medical Imaging, Western University, London, ON N6A 3K7, Canada. W. Xue and S. Li are also with the Digital Imaging Group of London, London, ON N6A 3K7, Canada.}%
\thanks{* Corresponding author. (E-mail: slishuo@gmail.com)}%
}

%
%

%

\maketitle

\begin{abstract}

Cardiac indices estimation is of great importance during identification and diagnosis of cardiac disease in clinical routine. However, estimation of multitype cardiac indices with consistently reliable and high accuracy is still a great challenge due to the high variability of cardiac structures and complexity of temporal dynamics in cardiac MR sequences. 
While efforts have been devoted into cardiac volumes estimation through feature engineering followed by a independent regression model, these methods suffer from the vulnerable feature representation and incompatible regression model.   
In this paper, we propose a semi-automated method for multitype cardiac indices estimation. After manual labelling of two landmarks for ROI cropping, an integrated deep neural network Indices-Net is designed to jointly learn the representation and regression models. It comprises two tightly-coupled networks: a deep convolution autoencoder (DCAE) for cardiac image representation, and a multiple output convolution neural network (CNN) for indices regression. Joint learning of the two networks effectively enhances the expressiveness of image representation with respect to cardiac indices, and the compatibility between image representation and indices regression, thus leading to accurate and reliable estimations for all the cardiac indices.

When applied with five-fold cross validation on MR images of 145 subjects, Indices-Net achieves consistently low estimation error for LV wall thicknesses (1.44$\pm$0.71mm) and areas of cavity and myocardium (204$\pm$133mm$^2$). It outperforms, with significant error reductions, segmentation method (55.1\% and 17.4\%) and two-phase direct volume-only methods (12.7\% and 14.6\%) for wall thicknesses and areas, respectively. These advantages endow the proposed method a great potential in clinical cardiac function assessment. 

\end{abstract}


%
\IEEEpeerreviewmaketitle

\section{Introduction}

\IEEEPARstart{C}{ardiac} disease is one of the leading cause of morbidity and mortality around the world. Accurate estimation of cardiac indices from cardiac MR images plays a critical role in early diagnosis and identification of cardiac disease. 
Cardiac indices are quantitative anatomical or functional information (wall thickness, cavity area, myocardium area, and ejection fraction (EF), etc.) of the heart, which help distinguish between pathology and health~\cite{peng2016review}. Two categories of solutions exist for cardiac indices estimation: traditional methods~\cite{ayed2009embedding, ayed2012max, peng2016review, petitjean2011review} and direct methods~\cite{wang2015prediction, afshin2012global, afshin2014regional, wang2014direct, zhen2014direct, zhen2015multi, zhen2015direct}. Traditional methods rely on the premise of cardiac segmentation, from which cardiac indices are then manually measured. However, obtaining good and robust segmentation is still a great challenge due to the diverse structure and complicate temporal dynamics of cardiac sequences, therefore resulting requirements of strong prior information and user interaction~\cite{ayed2009embedding, ayed2012max, peng2016review, petitjean2011review}. To circumvent these limitations, direct methods without segmentation have grown in popularity in cardiac volumes estimation, and obtained effective performance benefiting from machine learning algorithms ~\cite{wang2015prediction, afshin2012global, afshin2014regional, wang2014direct, zhen2014direct, zhen2015multi, zhen2015direct}. 

\begin{figure}[t]
	\centering
	\includegraphics[width=8cm]{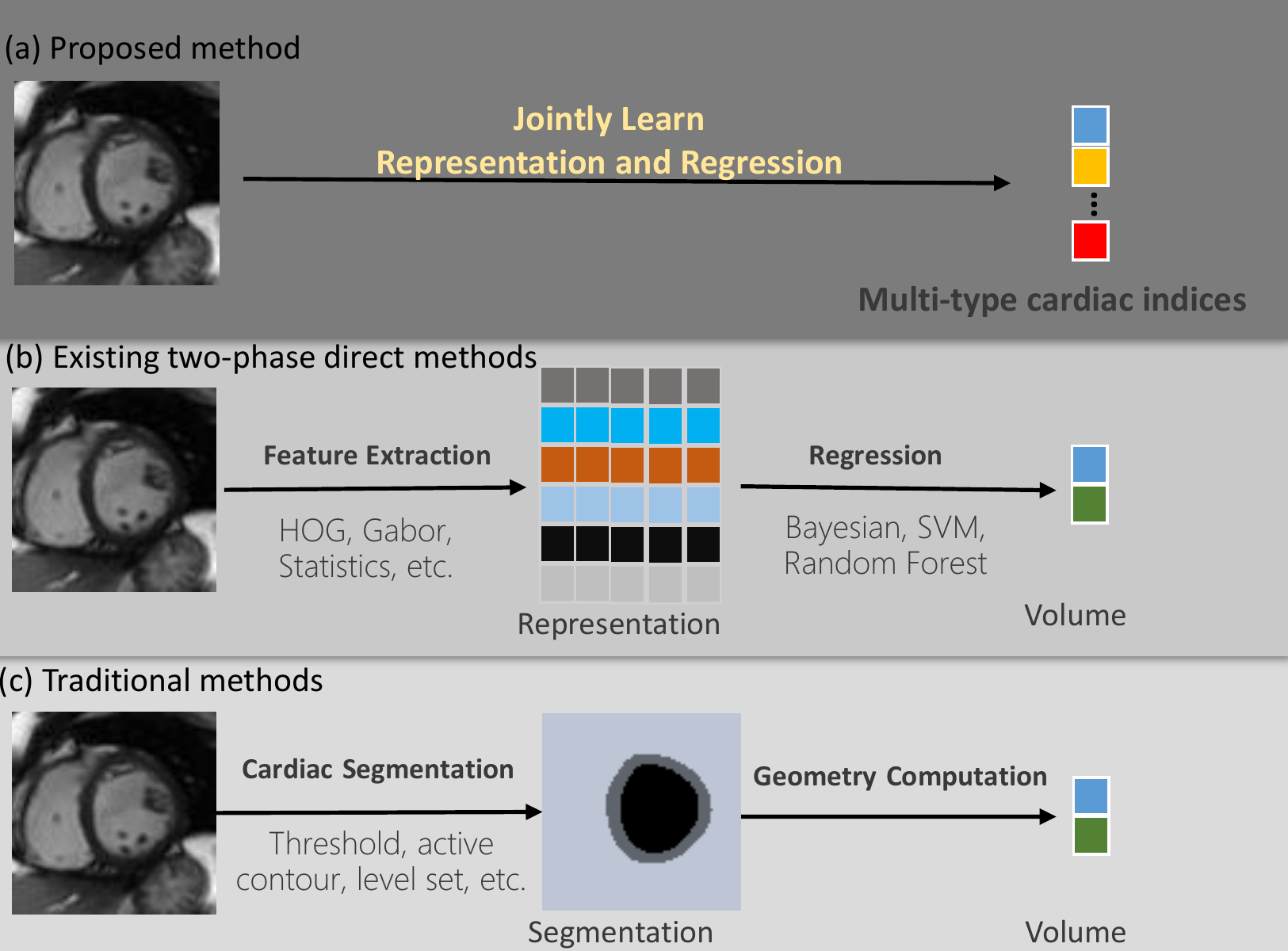}
	\caption{(a) The proposed method features advantages of multitype cardiac indices estimation and joint learning of image representation and the regression; (b) Existing direct methods predict only cardiac volumes and comprise two separately-handled phases, task-unaware feature extraction and target regression, which cannot maximumly benefit from each other; (c) Traditional segmentation-based methods compute indices from the segmented result which requires strong prior information and user interaction.}
	\label{fig_seg_direct}
\end{figure}

Despite their effective performance for volume estimation, existing direct methods can not obtain satisfactory results for multitype cardiac indices estimation. 1) They are not designed for \emph{multitype cardiac indices} estimation. Existing direct methods focus on estimation of volumes of ventricles and atriums only, whereas cardiac indices are far more than these volumes~\cite{peng2016review}. For other indices, such as wall thickness and myocardium area, more challenges compared to volume estimation arise (see Section~\ref{sub_sec_multiple}). 2) They do not learn image representation and regression models jointly and cannot make them adapt to and benefit from each other maximally. Existing direct methods follow a two-phase framework, where image representation is usually based on task-unaware hand-crafted features and the regression model is then learned separately, therefore obtain only inferior performance. 

To provide an accurate and reliable solution for multitype cardiac indices estimation, we propose a semi-automated method based on an integrated deep neural network \emph{Indices-Net}. It comprises two tightly-coupled networks: a deep convolution autoencoder (DCAE) for cardiac image representation, and a multiple output convolution neural network (CNN) for indices regression. When DCAE and CNN are learned jointly with proper initialization, Indices-Net can 1) remarkably enhance the expressiveness of image representation and the compatibility between image representation and indices regression models; and 2) simultaneously estimate with high accuracy two types of cardiac indices, i.e., linear indices (6 regional myocardium wall thicknesses), and planar indices (areas of LV myocardium and cavity). Experiments on 145 subjects show that Indices-Net achieves the lowest estimation error for linear indices (1.44$\pm$0.71mm) and planar indices (204$\pm$133mm$^2$). Fig.~\ref{fig_seg_direct} demonstrates how the proposed method differs from segmentation-based methods and existing two-phase volume-only direct methods.

\subsection{Multitype cardiac indices}\label{sub_sec_multiple}
\begin{figure}[t]
	\centering
	\subfloat{\includegraphics[width=3.5cm,height=3.5cm]{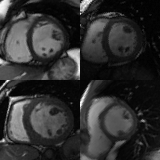}} 
	\hspace{3mm}
	\subfloat{\includegraphics[width=3.5cm,height=3.5cm]{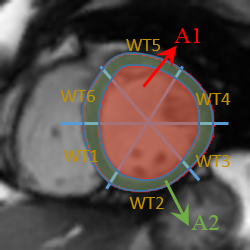}}
	
	\caption{Left: Variation of shape, contrast, density and presence of trabeculae and papillary muscles in cardiac MR images pose great challenge for  estimation of the cardiac indices. Right: the cardiac indices to be estimated in this work, which include area of LV cavity (A1), area of myocardium (A2), and the 6 regional wall thickness (WT1$\sim$WT6).}
    \label{fig_LV}
\end{figure}

Two types of cardiac indices~\cite{peng2016review} that describe anatomical information are to be estimated: linear indices (i.e, regional wall thickness of LV myocardium) and planar indices (i.e, areas of LV cavity and myocardium), as demonstrated in the right of Fig.~\ref{fig_LV}. These indices are closely related to regional and global cardiac function assessment, respectively. Detailed definitions and clinical roles of more cardiac indices can be found in \emph{Indices of cardiac function} of~\cite{peng2016review}. Despite their clinical significance~\cite{kawel2012normal, puntmann2013left}, multitype cardiac indices estimation has never been explored in existing direct methods~\cite{afshin2012global, kabani2016estimating, wang2014direct, zhen2014direct, zhen2015multi, zhen2015direct}. These methods only focus on volume estimation, which can be simplified as integration of cavity area along the long axis and is less difficult to estimate due to the high contrast of the boundary between LV cavity and myocardium, and the high density of the cavity area in cardiac MR images.   

More challenges arise during estimation of the above mentioned multitype cardiac indices. 1) Linear indices differ from planar ones in their relation with 2D spatial image structures, therefore more relevant and robust image representation is required to estimate both of them.  
2) As for the specific indices, regional wall thicknesses and myocardium area are susceptible to the complicated dynamic deformation of myocardium during the cardiac cycle, and the invisible epicardium boundary near the lateral free wall. Regional wall thicknesses are also subject to the orientation of myocardium segments in different regions. The representation and regression models should be capable of tolerating the dynamic deformation, the imperceptible boundary and the orientation variation.  

\subsection{Existing two-phase direct methods}

Two-phase framework employed in existing direct methods~\cite{afshin2012global, afshin2014regional, wang2014direct, zhen2014direct, zhen2015multi, zhen2015direct} is inadequate to achieve accurate estimation for multitype cardiac indices, for the reason that image representation and indices regression are separately handled, and no feedback connection exists between them during optimization. In the work of~\cite{afshin2012global}, LV cavity area estimation was conducted through feeding directly to the neural network the proposed image statistics, which were based on the Bhattacharyya coefficient between image distributions. Avoiding the requirement of segmentation, this method still needs user interaction to indicate two closed curves within and enclosing the cavity. The proposed statistical features were further employed in the detection of regional LV abnormalities~\cite{afshin2014regional}, where manual segmentation of a reference frame was required. In the work of~\cite{wang2014direct}, a Bayesian framework was build for bi-ventricular volume estimation based on multiple appearance features such as blob, homogeneity, and edge. Besides its intensive computation burden, an over simple linear correlation between areas of the two ventricular was taken into account in the prior model. Another direct estimation of bi-ventricular volumes was proposed in~\cite{zhen2014direct} with low level image features, i.e, Gabor features, HOG, and intensity, as input and random forest (RF)~\cite{breiman2001random} for feature selection and regression. These handcrafted features were further replaced with a more effective image representation that learned from a multiscale convolutional deep belief network (MCDBN)~\cite{zhen2015multi}, leading to improved correlation between the estimated volumes and their ground truth. Supervised descriptor learning (SDL) was proposed in the work of four chamber volumes estimation~\cite{zhen2015direct}, which still employed a separate adaptive K-clustering random forest (AKRF) regression~\cite{hara2014growing}. The compatibility between the descriptor and the regression model still can not be enhanced in the two-phase framework. 

\subsection{Deep neural network}
Deep neural networks have demonstrated great power in a broad range of visual applications, as well as medical image analysis~\cite{wang2016deep,fritscher2016deep} for the capability of learning effective hierarchy representations in an end-to-end fashion~\cite{zeiler2014visualizing}. Image representation obtained in such a way is endowed with more expressiveness with respect to the manifold structures of the image space and the target space. This property makes deep network quite suitable for the problem of cardiac indices estimation. However, in the area of cardiac image, deep networks are mostly deployed in segmentation with dense supervision of manually segmented results. Deep belief network, stacked autoencoder, and convolution neural network have been employed in cardiac segmentation with optional refinement by traditional models~\cite{ngo2013left,ngo2016combining,avendi2016combined,tran2016fully}. Fully convolution network (FCN) was applied to cardiac segmentation~\cite{tran2016fully} due to its success in semantic segmentation of natural image~\cite{FCN}. Recurrent FCN was later proposed to leverage inter-slice spatial dependencies in cardiac segmentation~\cite{poudel2016recurrent}. 

Only one work for cardiac volume estimation~\cite{kabani2016estimating} was proposed leveraging the hierarchy representation of deep neural network. In this work, a volume estimation CNN network took as input end-systole and end-diastole cardiac images (chosen by thresholding) of all slice positions, and output only two volume estimations for frames of end-systole and end-diastole. This makes the learning procedure more data-demanding and the network incapable of giving frame-wise prediction for detailed and reliable cardiac function assessment. Our proposed method is capable of estimating multitype cardiac indices for all frames throughout the whole cardiac cycle.
	
\vspace{0.5\baselineskip}
The remainder of this paper is organized as follows. Section II describes in detail the proposed Indices-Net, including the joint learning procedure, the representation network DCAE and the regression network CNN. Section III gives detailed descriptions of dataset, configuration, evaluation, and experiments. The results are reported and analyzed in Section IV. Conclusions and discussions are given in Section V.

\section{Methodology}

\begin{figure}[h]
	\centering
	\includegraphics[width=8.5cm]{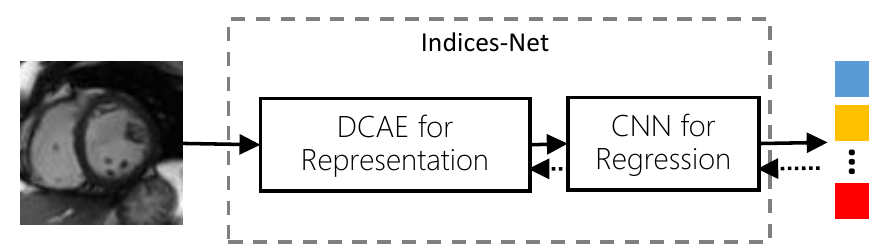}
	\caption{Joint learning of representation and regression model for multitype cardiac indices estimation by Indices-Net. Two tightly-coupled networks are included: DCAE for image representation and CNN for multiple indices regression. The two parts are learned with iterated forward propagation (solid arrows) and backward propagation (dashed arrows) to maximally benefit each other.}
	\label{fig_twophase}
\end{figure}

We propose an integrated deep network Indices-Net (Fig.~\ref{fig_twophase}) to jointly learn the image representation and regression for multitype cardiac indices estimation. A deep convolution autoencoder DCAE (19 layers) is designed to extract common expressive representation for all the indices, leveraging the capability of extracting discriminative feature and reconstructing the image in a generative manner. A shallow network CNN (3 layers) is tightly coupled to DCAE, to further extract index-specific feature and predict the corresponding cardiac index. Joint learning of DCAE and CNN remarkably enhances the expressiveness of image representation and the compatibility between image representation and the regression, therefore leads to highly reliable and accurate estimations. The learning procedure and details of the two networks are given below.

\subsection{Joint learning of representation and regression}

Joint learning intends to improve the expressiveness of image representation with supervision of cardiac indices, and alleviates the demanding of complicated regression model. In our work, joint learning of DCAE and CNN is implemented by iterated forward/backward propagation of the integrated whole network. During forward propagation (solid arrows in Fig.~\ref{fig_twophase}), DCAE extracts common image representations from cardiac MR images, and then CNN disentangles from this representation index-specific features for each index, from which the indices are predicted. During backward propagation (dashed arrows in Fig.~\ref{fig_twophase}), the estimation error of cardiac indices is then back-propagated to update the parameters in CNN, and further backward through all layers of DCAE, to update the common representation for cardiac images and embed indices information into this representation. Iteration of this alternative forward/backward propagation constantly enhances the expressiveness of the obtained image representation with respect to cardiac indices, as well as its compatibility with the regression network. Once the learning procedure is finished, one-pass forward propagation is sufficient to achieve estimations of cardiac indices for novel images.  

\subsection{Cardiac image representation learning with DCAE}\label{sec_dcae}

In this work, the representation network is designed as a customized deep convolution autoencoder (DCAE), with input dimension and filter numbers for each layer accommodated to the single-channel gray cardiac images. DCAE is composed of two subparts: convolution layers which are capable of extracting latent discriminative structural features from the input cardiac images; and deconvolution layers which are capable of reconstructing the output from these latent features in a generative manner. With both of them, DCAE is capable of building a cascade mapping of \emph{cardiac image (input layer)$\rightarrow$ latent representation (fc6)$\rightarrow$ indices-relevant structure} by the bottom-up discriminative encoder (convolution layers) and the top-down generative decoder (deconvolution layers) together. In such a way, expressive representations with respect to the cardiac indices can be obtained.

Convolution autoencoder (CAE) was firstly introduced in~\cite{masci2011stacked} by two convolution layers with their weight matrices being transposed of each other. A stack of CAE was trained to initialize a CNN-based classifier. The convolution layer with transposed weight matrix is later replaced by a deconvolution layer whose weights are learnable~\cite{noh2015learningdeconv}. The deconvolution layer was proposed in~\cite{zeiler2010deconvolutional} to build low and mid-level image representation in a generative manner and is closely related to convolutional sparse coding~\cite{bristow2013fast}. The learned filters act as the basis in sparse coding and are capable of capturing rich structures of different types.

\begin{figure}[t]
	\centering
	\subfloat[Architecture of DCAE, which constitutes two mirrored subparts: the discriminitive convolution layers and the generative deconvolution layers. With both of them, a mapping between the input and the output of DCAE is built.]{\includegraphics[width=8.5cm]{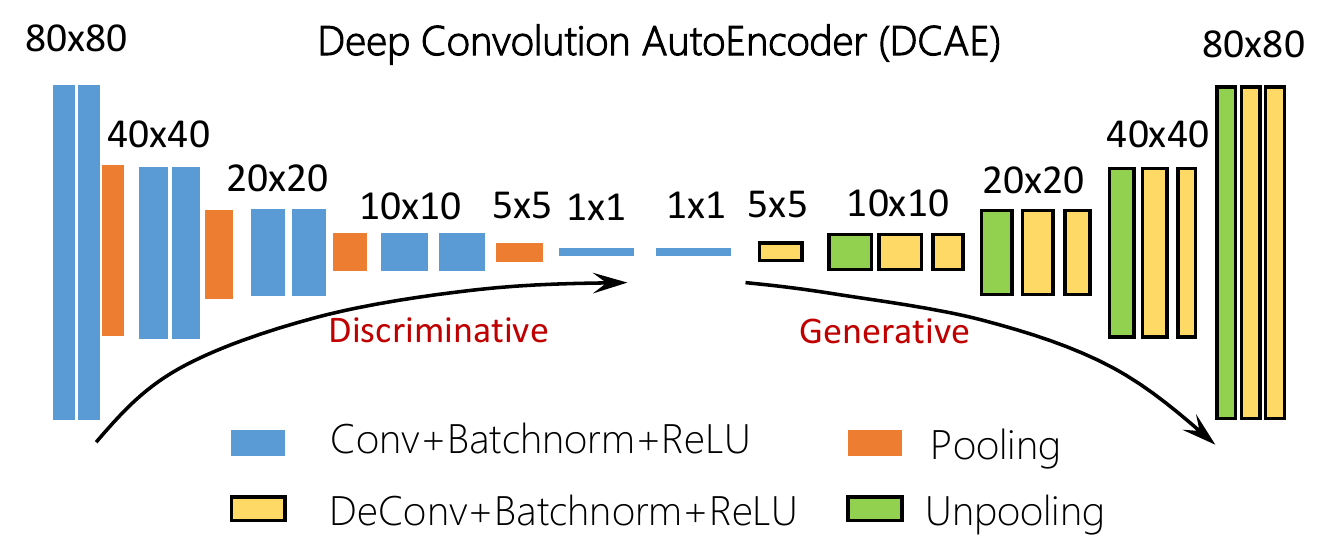}}
	\hfil
	\subfloat[Pretraining of DCAE as a cardiac image autoencoder.]{\includegraphics[width=8cm]{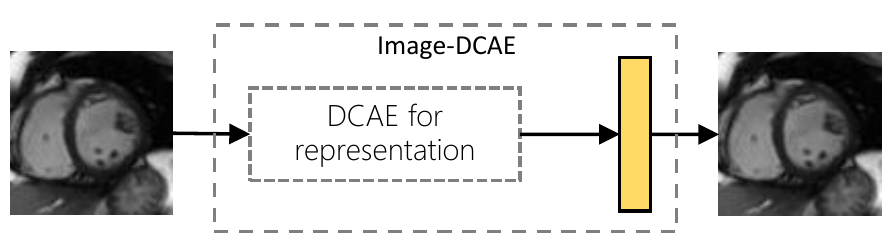}}
	\caption{Architecture of DCAE and pre-training of DCAE with cardiac images.}
	\label{fig_deconvnet}
\end{figure}

The architecture of our DCAE is shown in Fig.~\ref{fig_deconvnet}(a). The subpart of convolution has 9 convolution layers, with batch normalization layer and ReLU layer following each of them. Batch normalization layer~\cite{ioffe2015batch} helps reduce internal covariance shift in very deep network and accelerate the network convergence. Max pooling is performed every two convolution layers. The subpart of deconvolution is a mirrored version of the convolution subpart, with convolution layer and pooling layer replaced by deconvolution layer and unpooling layer. A fully connected layer in between connects the two parts, transforming the features obtained from each part. 

In DCAE, two types of layers are important: deconvolution layer and unpooling layer.

\begin{figure}[t]
	\centering
	\subfloat[unpool3, $20\times20$]{\includegraphics[width=4cm,trim={1.9cm 0.59cm 1.4cm 0.35cm},clip]{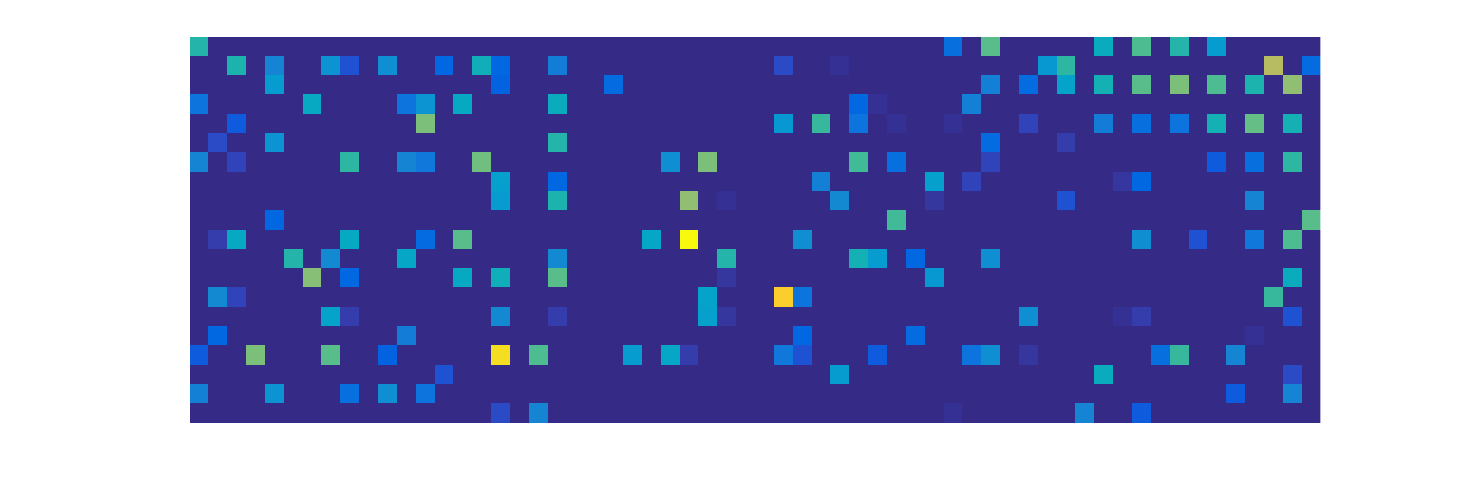}}
	\hfill
	\subfloat[deconv3-2, $20\times20$]{\includegraphics[width=4cm,trim={1.9cm 0.59cm 1.4cm 0.35cm},clip]{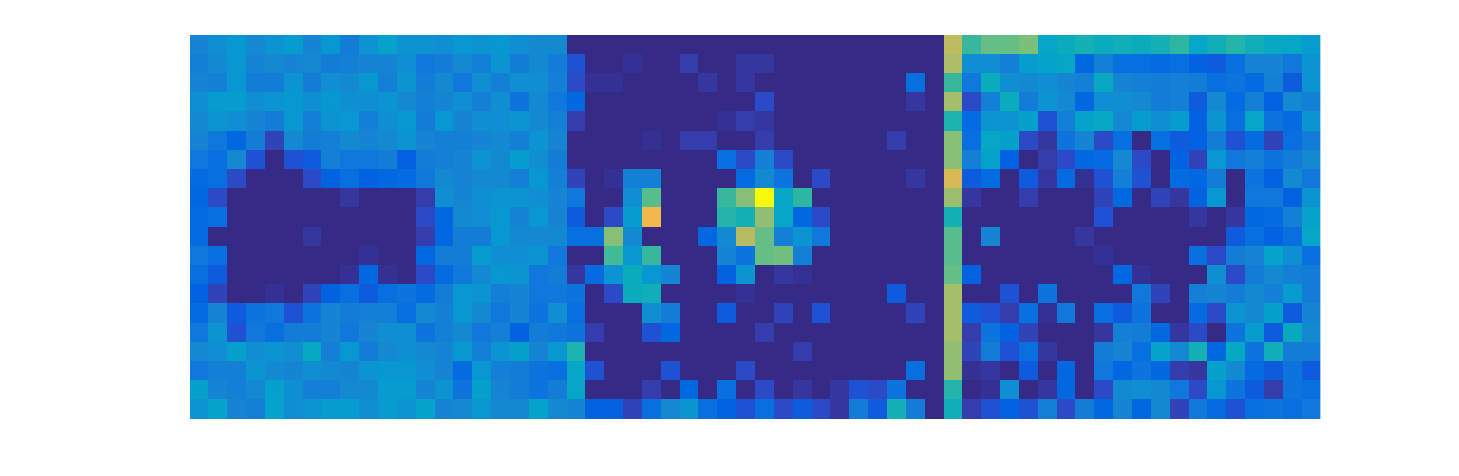}}
	\hspace{0mm}
	\subfloat[unpool2, $40\times40$]{\includegraphics[width=4cm,trim={1.9cm 0.59cm 1.4cm 0.35cm},clip]{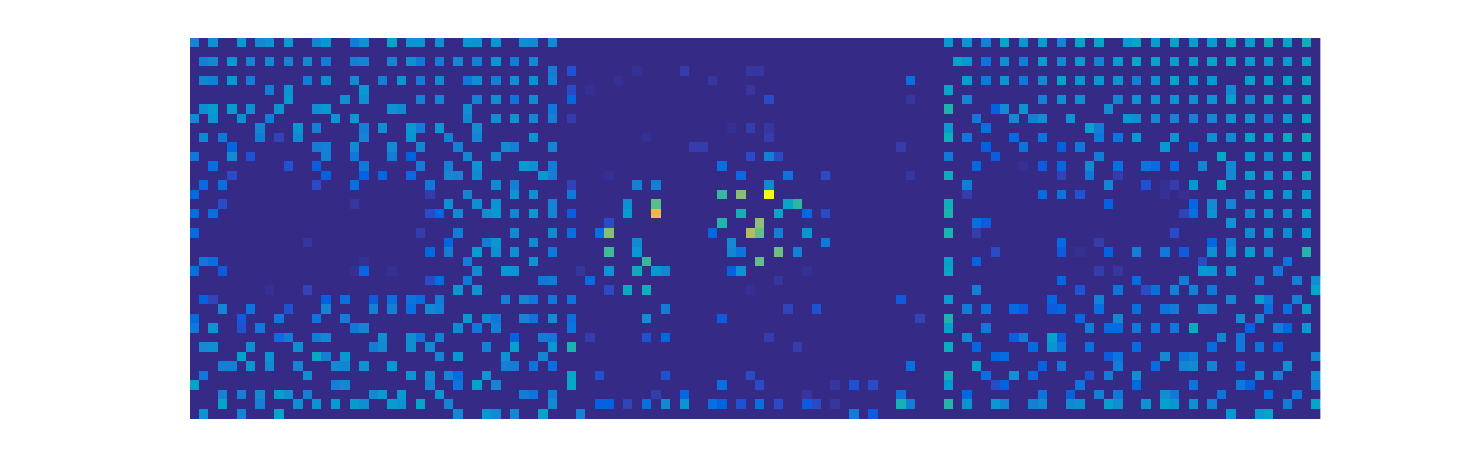}}
	\hfill
	\subfloat[deconv2-2, $40\times40$]{\includegraphics[width=4cm,trim={1.9cm 0.59cm 1.4cm 0.35cm},clip]{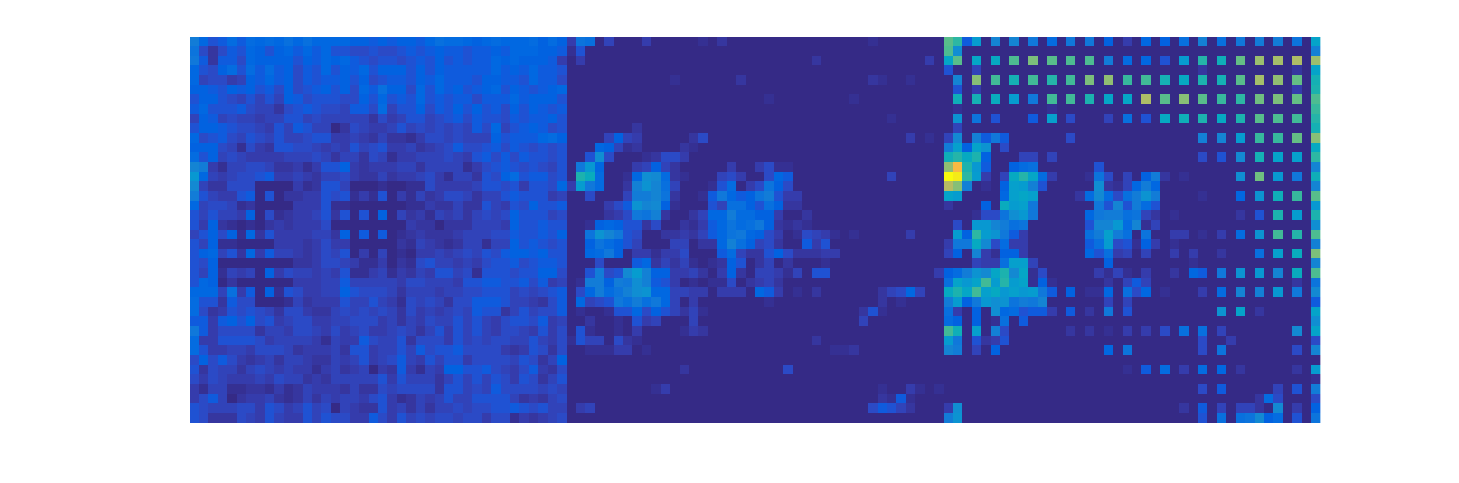}}
	\hspace{0mm}
	\subfloat[unpool1, $80\times80$]{ \includegraphics[width=4cm,trim={1.9cm 0.59cm 1.4cm 0.35cm},clip]{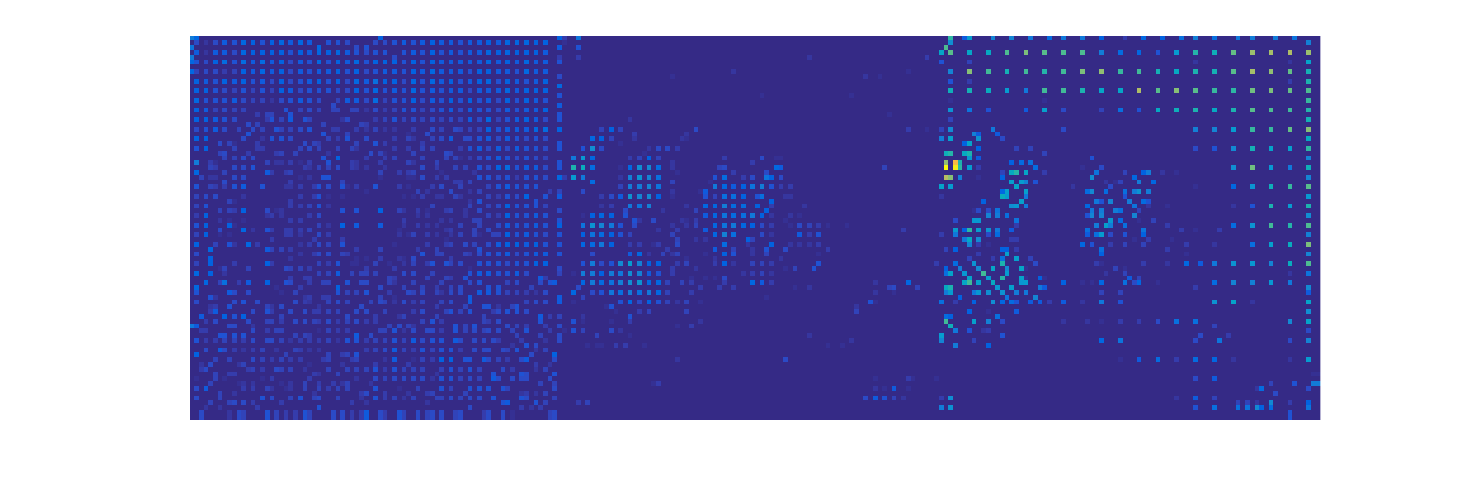}}
	\hfill
	\subfloat[deconv1-2, $80\times80$]{\includegraphics[width=4cm,trim={1.9cm 0.59cm 1.4cm 0.35cm},clip]{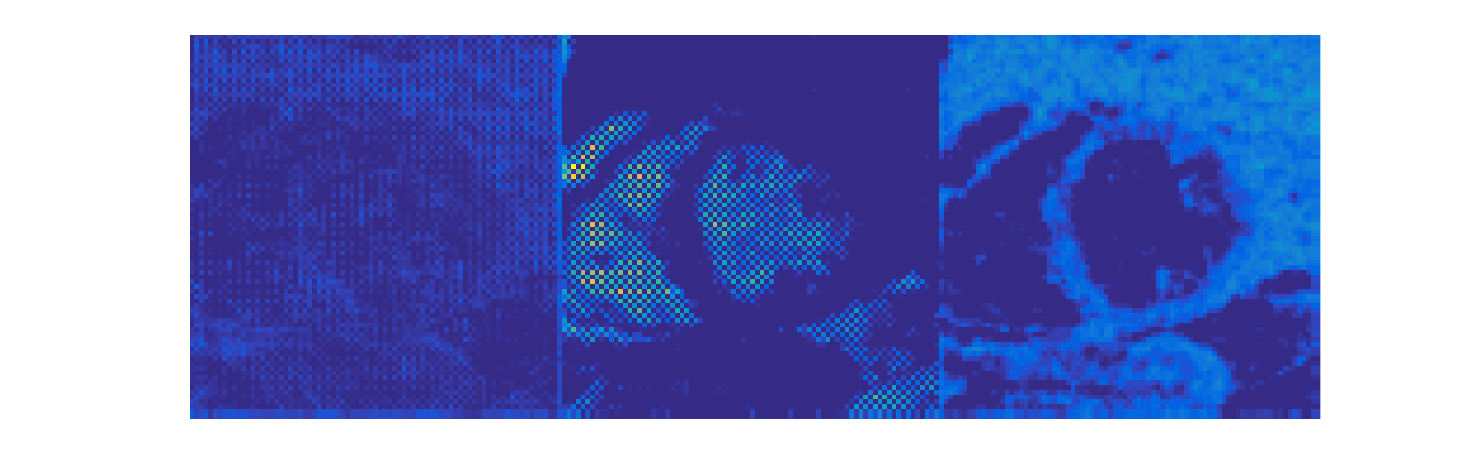}}
	\hspace{0mm}
	\subfloat[input]{\includegraphics[width=2.3cm]{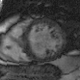}}
	\hspace{20mm}
	\subfloat[reconstruction]{\includegraphics[width=2.3cm]{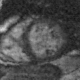}}
	\caption{Visualization of the feature maps generated by Image-DCAE. For each considered layer (refer Table.~\ref{table_config} for layer's name), three representative feature maps for the input image (g) are illustrated. Each of the feature maps favors some specific structures in the cardiac image such as LV cavity, background, and the myocardium.}
	\label{fig_featuremap}
\end{figure}
\begin{table}[t]
	\caption{Configuration of the network.}
	\label{table_config}
	\centering
	\begin{tabular}{c|ccc|c}
		\hline
		name & kernel size & stride & pad &output size\\
		\hline
		input & -&-&-& $80\times80\times 1$\\
		\hline
		\multicolumn{5}{c}{DCAE}\\
		\hline
		conv1-1& $3\times 3$&1&1&$80\times80\times 16$\\
		conv1-2& $3\times 3$&1&1&$80\times80\times 16$\\
		pool1  & $2\times 2$&2&0&$40\times40\times 16$\\
		\hline
		conv2-1& $3\times 3$&1&1&$40\times40\times 32$\\
		conv2-2& $3\times 3$&1&1&$40\times40\times 32$\\
		pool2  & $2\times 2$&2&0&$20\times20\times 32$\\
		\hline
		conv3-1& $3\times 3$&1&1&$20\times20\times 64$\\
		conv3-2& $3\times 3$&1&1&$20\times20\times 64$\\
		pool1  & $2\times 2$&2&0&$10\times10\times 64$\\
		\hline
		conv4-1& $3\times 3$&1&1&$10\times10\times 128$\\
		conv4-2& $3\times 3$&1&1&$10\times10\times 128$\\
		pool4  & $2\times 2$&2&0&$5\times5\times 128$\\
		\hline
		conv5    & $5\times 5$&1&0&$1\times 1\times 512$\\
		\hline
		fc6    & $1\times 1$&1&0&$1\times 1\times 512$\\
		\hline
		deconv5& $5\times 5$&1&0&$5\times 5\times 128$\\
		\hline
		unpool4& $2\times 2$&2&0&$10 \times 10\times128$\\
		deconv4-1&$3\times 3$&1&1&$10 \times 10\times 128$\\
		deconv4-2&$3\times 3$&1&1&$10\times 10\times 64$\\
		\hline
		unpool3& $2\times 2$&2&0&$20 \times 20\times64$\\
		deconv3-1&$3\times 3$&1&1&$20 \times 20\times 64$\\
		deconv3-2&$3\times 3$&1&1&$20\times 20\times 32$\\
		\hline
		unpool2& $2\times 2$&2&0&$40 \times 40\times32$\\
		deconv2-1&$3\times 3$&1&1&$40 \times 40\times 32$\\
		deconv2-2&$3\times 3$&1&1&$40\times 40\times 16$\\
		\hline
		unpool1& $2\times 2$&2&0&$80 \times 80\times16$\\
		deconv1-1&$3\times 3$&1&1&$80 \times 80\times 16$\\
		deconv1-2&$3\times 3$&1&1&$80\times 80\times 16$\\
		\hline
		\multicolumn{5}{c}{Image-DCAE}\\
		\hline
		dconv-rec&$1\times 1$&1&0&$80\times80\times 1$\\
		\hline
		\multicolumn{5}{c}{Indices-Net}\\
		\hline
		conv-reg1&$5\times 5$&1&2&$80\times 80\times 16$\\
		conv-reg2&$5\times 5$&1&2&$80\times 80\times 16$\\
		conv-reg3&$80\times 80$&1&0&$1\times 1\times 8$\\
		\hline
	\end{tabular}
\end{table}

\subsubsection{Deconvolution}
Denoted by $z_{l}\in R^{n_l\times n_l\times c_{l}}$ the output of layer $l$ with $c_l$ channels, each associated with kernel $h_{l,k}\in R^{m\times m\times c_{l-1}}, k=1,...c_l$, the convolution layer can be described as:   
\begin{equation}
z_{l,k} = f(z_{l-1}\ast h_{l,k}), ~k=1,...c_l.
\end{equation}
$f$ denotes the element-wise nonlinear transformation of ReLU and batch normalization. 
Define $\mathcal{R}_m(x)$ a operation which extracts all the patches of size ${m\times m}$ in $x$ along the two spatial dimensions and rearrange them into columns of a matrix, the convolution layer with all output channels can be reformulated as:
\begin{equation}\label{eq_conv}
\begin{split}
&\mathbf{z}_{l}=f(\mathbf{h}_{l}^T\mathbf{z}_{l-1})\\
&with~~ \mathbf{z}_{l-1}=\mathcal{R}_m(z_{l-1}), \mathbf{h}_{l}=\mathcal{R}_m(h_{l}), \mathbf{z}_{l}=\mathcal{R}_1(z_{l})
\end{split}
\end{equation} 
$\mathbf{z}_{l-1} \in R^{m^2c_{l-1}\times n_{l-1}^2}$ is the matrix form of the input, and $\mathbf{h}_{l}\in R^{m^2c_{l-1}\times c_l}$ maps multiple inputs within a receptive field to one single output in $\mathbf{z}_{l}\in R^{c_l\times n_l^2}$. 

The deconvolution layer, on the contrary, reverses the convolution operation by associating a single activation in its input with multiple activations in the output: 
\begin{equation}\label{eq_deconv}
\begin{split}
&\mathbf{z}_{l}=f(\mathbf{h}_{l}^T\mathbf{z}_{l-1})\\
&with~~\mathbf{z}_{l-1}=\mathcal{R}_1(z_{l-1}), \mathbf{h}_{l}=\mathcal{R}_1(h_{l}), \mathbf{z}_{l}=\mathcal{R}_m(z_{l})
\end{split}
\end{equation}
where $\mathbf{z}_{l-1}\in R^{c_{l-1}\times n_{l-1}^2}$, $\mathbf{h}_{l}\in R^{c_{l-1}\times m^2c_l}$, and $\mathbf{z}_{l}\in R^{m^2c_l\times n_l^2}$. The output of a deconvolution layer can then be obtained by $z_{l}=\mathcal{R}_m^{-1}(\mathbf{z}_{l})$. Note that the difference of Eq.~\ref{eq_conv} and~\ref{eq_deconv} lies in the patch size of the operation $\mathcal{R}_m$. The kernel of a deconvolution layer is updated in the same way as in a convolution layer independently.

\subsubsection{Unpooling}
Another important layer in DCAE is the unpooling layer, which reverses the corresponding pooling layer with respect to the switches of the max pooling operation. For a max pooling operation $P_m$ with kernel size $m\times m$, the output is:
\begin{equation}
[p_{l,k},s_{l,k}] = P_m(z_{l,k})
\end{equation}
where the pooled maps $p_{l,k}$ store the values and switches $s_{l,k}$ record the locations. The unpooling operation $U_m$ takes elements in $p_{l,k}$ and places them in $\hat{z}_{l,k}$ at the locations specified by $s_{l,k}$. 
\begin{equation}
\hat{z}_{l,k}=U_m(p_{l,k},s_{l,k})
\end{equation} 

Unpooling layer is particularly useful to reconstruct the structure of input image by tracing back to image space the locations of strong activations. Details of deconvolution layer and unpooling layer can be found in~\cite{noh2015learningdeconv,zeiler2010deconvolutional,zeiler2014visualizing}.

\vspace{0.5\baselineskip}
\emph{Initialization with pre-trained DCAE}

To alleviate the training procedure of the whole network and equip DCAE with expressiveness of cardiac image structure, DCAE is initialized with parameters pre-trained from a cardiac image autoencoder Image-DCAE. Unsupervised pre-training has been proved to behave as a form of regularization towards the parameter space and support better generalization~\cite{erhan2010does}. In our network, Image-DCAE is constructed by adding a deconvolution layer with one output channel on top of the DCAE to reconstruct the input image, as shown in Fig.~\ref{fig_deconvnet}(b). After being trained with cardiac MR images (no label is required here), Image-DCAE is capable of extracting different abstract levels of structures in cardiac images. Fig.~\ref{fig_featuremap} shows the feature maps (a-f) of a cardiac image (g) and its reconstructed result (h) generated by Image-DCAE. For each considered layer, we show three representative feature maps. These feature maps favor some specific structures in the cardiac image that are responsive to cardiac indices considered in this work, such as LV cavity, background, and the myocardium. As these feature maps are forward-propagated to higher deconvolution layers, finer details of the cardiac structure can be revealed. With these parameters capturing cardiac structure as initialization, it becomes more efficient to obtain indices relevant information during the iterated joint learning procedure. Benefits of this initialization can also be found in Section~\ref{seg_benefit_pretrain}. 

\subsection{Multitype cardiac indices estimation with CNN}\label{seg_cnn}

\begin{figure}[h]
	\centering
	\includegraphics[width=8.5cm]{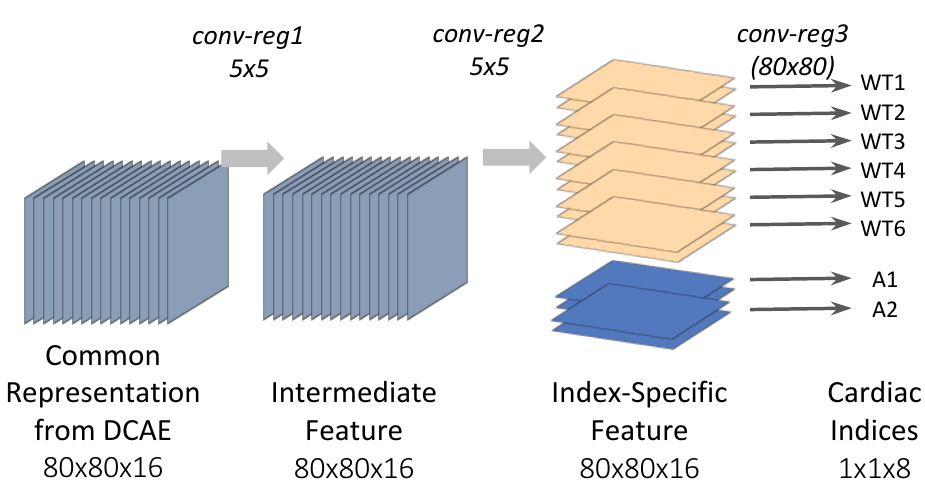}
	\caption{Index-specific feature extraction (first two layers) and regression (third layer) for multiple cardiac indices estimation.}
	\label{fig_reg_net}
\end{figure}

\begin{figure*}[t]
	\centering
	\includegraphics[width=13cm,trim={0cm 1.5cm 0cm 0cm},clip]{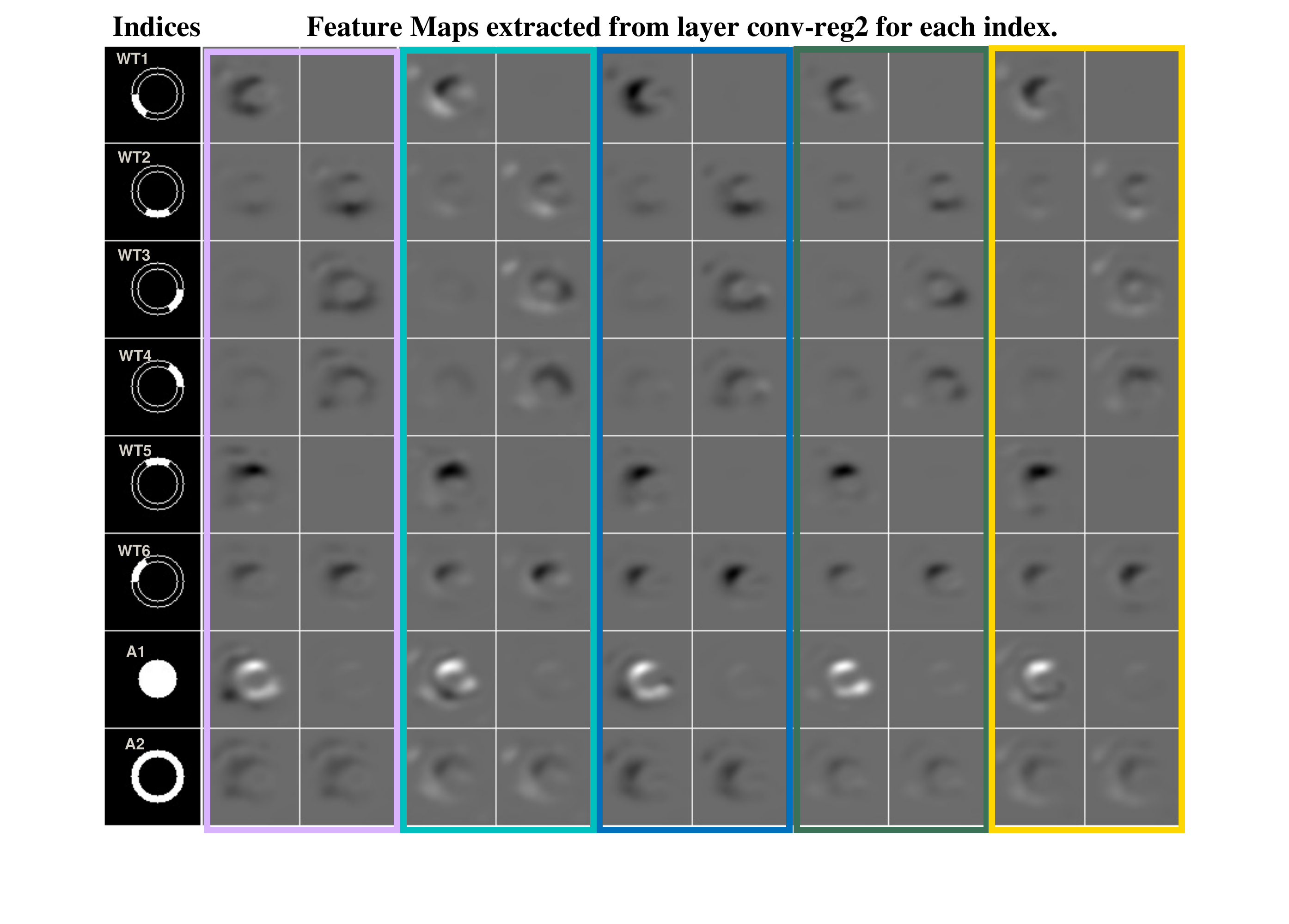}
	\caption{Indices-specific feature maps extracted with layer conv-reg2 for 5 cardiac images (shown in each color rectangle). The diagrams in the first column indicate the corresponding indices and their spatial relations. In each row show the results for one cardiac index indicated by the leftmost diagram. The dark/bright regions in these feature maps approximately correspond to the most responsive features for the index. Overlap of these feature maps between different indices captures their correlation.}
	\label{fig_regscore5}
\end{figure*}

To estimate multitype cardiac indices with image representation from DCAE, we design a simple CNN with index-wise connection on top of DCAE, as shown in Fig.~\ref{fig_reg_net}. We find that there is no need for a deeper or more complex regression network, since the representation network DCAE has extracted expressive information from cardiac image. With the outputs of DCAE as common representations, the first two layers in CNN aim at disentangling index-specific features from them, resulting two feature maps for each index, as shown in Fig.~\ref{fig_regscore5}. The third layer estimates each cardiac index from the corresponding feature maps with a simple linear model. Because common representations obtained by DCAE are employed here, the correlations among these indices are automatically embedded through overlapping of the extracted index-specific feature maps. 

Fig.~\ref{fig_regscore5} demonstrates for 5 cardiac images (within color rectangles) the index-specific features obtained with layer conv-reg2. In each row of the rectangle show two features maps for one cardiac index indicated by the diagrams in the leftmost column. As shown in the figure, Indices-Net can capture the most responsive features (the bright/dark regions) from cardiac images to estimate each index. The overlapping of feature maps for different indices alleviate the estimation of some challenging indices, such as WT3 and WT4, by leveraging the correlation between neighbouring indices.

\section{Experiments}

\subsection{Dataset}\label{sec_database}

To evaluate the performance of our method, a dataset of 2D short-axis cine MR images with labelled cardiac indices is used, which includes 2900 images from 145 subjects. These subjects are collected from 3 hospitals affiliated with two health care centers (London Healthcare Center and St. Joseph’s HealthCare)
using scanners of 2 vendors (GE and Siemens). The subjects age from 16 yrs to 97 yrs, with average of 58.9 yrs. The pixel spacings of the MR images range from 0.6836 mm/pixel to 2.0833 mm/pixel, with mode of 1.5625 mm/pixel. Diverse pathologies are in presence including regional wall motion abnormalities, myocardial hypertrophy, mildly enlarged LV, atrial septal defect, LV dysfunction, etc. Each subject contains 20 frames throughout a cardiac cycle. In each frame, LV is divided into equal thirds (basal, mid-cavity, and apical) perpendicular to the long axis of the heart following the standard AHA prescription~\cite{cerqueira2002standardized} and a representative mid-cavity slice is selected for validation of cardiac indices estimation.    

Several preprocessing steps are applied prior to ground truth calculation. 1) Landmark labelling. Two landmarks, i.e, junctions of the right ventricular wall with the left ventricular, are manually labelled. 2) Rotation. Each cardiac image is rotated until the line between the two landmarks is vertical. 3) ROI cropping. The ROI image is cropped as a squared region centered at the mid-perpendicular of this line and with size twice the distance between the two landmarks. 4) Resizing. All the cropped images are resized to the dimension of $80\times 80$.  

After the preprocessing, all the cardiac images are manually contoured to obtained the epicardial and endocardial borders, which are double-checked by two experienced cardiac radiologists (A. Islam and M. Bhaduri). The ground truth values of LV cavity area and myocardium area can be easily obtained by counting the pixel numbers in the segmented cavity and myocardium. The linear-type regional wall thicknesses are obtained as follows. First, myocardial thicknesses are automatically acquired from the two borders in 60 measurements using the 2D centerline method~\cite{buller1997assessment}. Then the myocardium is divided into 6 segments (as shown in Fig.4 of~\cite{cerqueira2002standardized}), with 10 measurements per segment. Finally, these measurements are averaged per segment as the ground truth of regional wall thicknesses. Papillary muscles and trabeculations are excluded in the myocardium. The obtained two types of cardiac indices are normalized according to the image dimension (80) and area (6400), respectively. During evaluation, the obtained results are converted to physical thickness (in mm) and area (in mm$^2$) by reversing the resizing procedure and multiplying the pixel spacing for each subject.

\begin{table*}[!ht]
	\caption{Performance comparison of the proposed method with segmentation based method and existing direct methods for wall thicknesses (mm) and areas of LV cavity and myocardium (mm$^2$). MAE and $\rho$ are illustrated in each cell. Best results are highlighted in bold for each column. (Note that the mode of pixel-spacing in our dataset of 1.5625mm/pixel.)}
	\label{table_compare_wt}
	\centering
	\setlength{\tabcolsep}{3pt}
	\begin{tabular}{*c||^c^c^c^c^c^c|^c||^c^c|^c}
		\hline
		\multirow{2}{*}{Method}&\multicolumn{7}{c||}{linear indices (mm)}&\multicolumn{3}{c}{planar indices (mm$^2$)}\\
		\cline{2-11}
		& WT1 & WT2& WT3& WT4& WT5& WT6& Average&A1&A2&Average\\
		\hline
		\multirow{2}{*}{Max Flow~\cite{ayed2012max}}&1.53$\pm$1.73&3.23$\pm$2.83  &4.15$\pm$3.17 &5.08$\pm$3.95  &3.47$\pm$3.25 &1.76$\pm$1.80  &3.21$\pm$1.98&\textbf{156$\pm$193} &339$\pm$272 &247$\pm$201\\
		&0.796&\textbf{0.720}&\textbf{0.743}&\textbf{0.706}&0.724&0.785&0.746&\textbf{0.958}&0.851&\textbf{0.904}\\
		\hline
		\multirow{2}{*}{Multi-features+RF~\cite{zhen2014direct}} & 1.70$\pm$1.47  &1.71$\pm$1.34  &1.97$\pm$1.54 &1.82$\pm$1.41 &1.55$\pm$1.33 &1.68$\pm$1.43&1.73$\pm$0.97&231$\pm$193 &291$\pm$246  &261$\pm$165\\
		&0.729&0.603&0.483&0.533&0.685&0.777&0.635&0.924&0.729&0.827\\
		\hline
		\multirow{2}{*}{SDL+AKRF~\cite{zhen2015direct}} &1.98$\pm$1.58 &1.67$\pm$1.40 &1.88$\pm$1.63 &1.87$\pm$1.55  &1.65$\pm$1.45 &2.04$\pm$1.59 &1.85$\pm$1.03&198$\pm$169&286$\pm$242&242$\pm$158\\
		&0.599&0.582&0.515&0.493&0.599&0.626&0.569&0.942&0.742&0.842\\
		\hline
		\multirow{2}{*}{MCDBN+RF~\cite{zhen2015multi}} &1.78$\pm$1.40 &1.68$\pm$1.41 &1.92$\pm$1.45 &1.66$\pm$1.20 &\textbf{1.20$\pm$1.01} & 1.63$\pm$1.23 &1.65$\pm$0.77  &208$\pm$166 &269$\pm$217 &239$\pm$135\\
		&0.611&0.462&0.435&0.547&0.661&0.726&0.573&0.926&0.723&0.824\\
		\hline
		\rowstyle{\bfseries}
		\multirow{2}{*}{Indices-Net}& 1.39$\pm$1.13 &1.51$\pm$1.21&1.65$\pm$1.36 &1.53$\pm$1.25 & \normalfont{1.30$\pm$1.12}&1.28$\pm$1.00 &1.44$\pm$0.71& \normalfont{185$\pm$162} & 223$\pm$193&  204$\pm$133 \\
		\rowstyle{\bfseries}
		&0.824&\normalfont{0.701}&\normalfont{0.671}&\normalfont{0.698}&0.781&0.871&0.758&\normalfont{0.953}&0.853&\normalfont{0.903}\\
		\hline
	\end{tabular}
\end{table*}

\subsection{Configurations}
In our experiments, 5-fold cross validation is employed for performance evaluation and comparison. The dataset is divided into 5 groups, each containing 29 subjects. Four groups are employed to train the prediction model, and the last group is used for test. This procedure is repeated five times until the indices of all subjects are obtained. The network is implemented by Caffe~\cite{jia2014caffe} with SGD solver. The configuration of the whole network Indices-Net is shown in Table.~\ref{table_config}. Learning rate and weight decay are set to (0.0001, 0.005) for Image-DCAE and (0.05, 0.02) for Indices-Net. In both procedures, \textquoteleft inv\textquoteright ~learning policy is employed with gamma and power being (0.001, 2) and momentum 0.9.

\subsection{Performance evaluation}
We first evaluate the estimation accuracy for the two types of cardiac indices with two criteria: correlation coefficient ($\rho$) and mean absolute error (MAE) between the estimated results and the ground truth. Denote  $\hat{y}_{s,f}^{ind}$ and $y_{s,f}^{ind}$ the estimated and ground truth cardiac index of the $s$th subject and the $f$th frame, where $ind\in\{WT1,...WT6, A1, A2\}$, $1\le s\le 145$, $1\le f\le 20$. The two criteria are calculated as follows:
\begin{equation}
MAE^{ind} = \frac{1}{145\times 20}\sum_{s=1}^{145}\sum_{f=1}^{20}|y_{s,f}^{ind}-\hat{y}_{s,f}^{ind}|
\end{equation}
\begin{equation}
\rho^{ind}=\frac{2\sum_{s=1}^{145}\sum_{f=1}^{20}(y_{s,f}^{ind}-y_{m}^{ind})(\hat{y}_{s,f}^{ind}-\hat{y}_{m}^{ind})}{\sum_{s=1}^{145}\sum_{f=1}^{20}((y_{s,f}^{ind}-y_{m}^{ind})^2+(\hat{y}_{s,f}^{ind}-\hat{y}_{m}^{ind})^2)}	
\end{equation}
where $y_{m}^{ind}$ and $\hat{y}_{m}^{ind}$ are the mean value of the ground truth and estimated indices $ind$.

To further evaluate the effectiveness of the estimated anatomical indices in cardiac function assessment, two cardiac functional indices are computed: Ejection Fraction, which quantifies the quantity of blood pumped out of the heart in each beat as percentage and indicates the global cardiac function; and Wall Thickening, which quantifies the change of myocardial wall thickness during systole as percentage and reflects regional cardiac function. For the $s$th subject, the two functional indices are computed as: 
\begin{equation}
\text{Ejection~Fraction}_s =\frac{y_{s,ED}^{A1}-y_{s,ES}^{A1}}{y_{s,ED}^{A1}}100\%
\end{equation}
\begin{equation}
\text{Wall~Thickening}_s=\frac{y_{s,ES}^{WT}-y_{s,ED}^{WT}}{y_{s,ED}^{WT}}100\%
\end{equation}
where ED and ES indicate end-diastole and end-systole frames, and the superscript ${WT}$ indicates the mean value of WT1$\sim$WT6. Correlation coefficients and MAE are computed for functional indices.

\begin{table}[t]
	\caption{Effectiveness of the proposed method for estimation of two functional cardiac indices: Ejection Fraction and Wall thickening. Results of competitors are included for comparison. MAE and $\rho$ are illustrated in each cell.}
	\label{table_compare_function}
	\centering
	\setlength{\tabcolsep}{3pt}
	\begin{tabular}{c||cc}
		\hline
		Method&Ejection Fraction&Wall Thickening\\
		\hline
		\multirow{2}{*}{Max Flow~\cite{ayed2012max}}&8.97$\pm$6.23\%&57.3$\pm$43.8\%  \\
		&\textbf{0.896}&0.604\\
		\hline
		\multirow{2}{*}{Multi-features+RF~\cite{zhen2014direct}} & 15.2$\pm$8.71\% &29.4$\pm$20.3\%\\
		&0.754&0.590\\
		\hline
		\multirow{2}{*}{SDL+AKRF~\cite{zhen2015direct}} &8.79$\pm$7.73\% &23.7$\pm$18.7\%\\
		&0.655&0.426\\
		\hline
		\multirow{2}{*}{MCDBN+RF~\cite{zhen2015multi}} &7.75$\pm$7.15\% &19.0$\pm$17.6\%\\
		&0.792&0.494\\
		\hline
		\multirow{2}{*}{Indices-Net}& \textbf{6.22$\pm$5.01\%} &\textbf{18.6$\pm$15.8\%}\\
		&0.856&\textbf{0.610}\\
		\hline
	\end{tabular}
\end{table}

\begin{figure*}[tbp]
	\centering
	\subfloat{\includegraphics[width=0.5\textwidth]{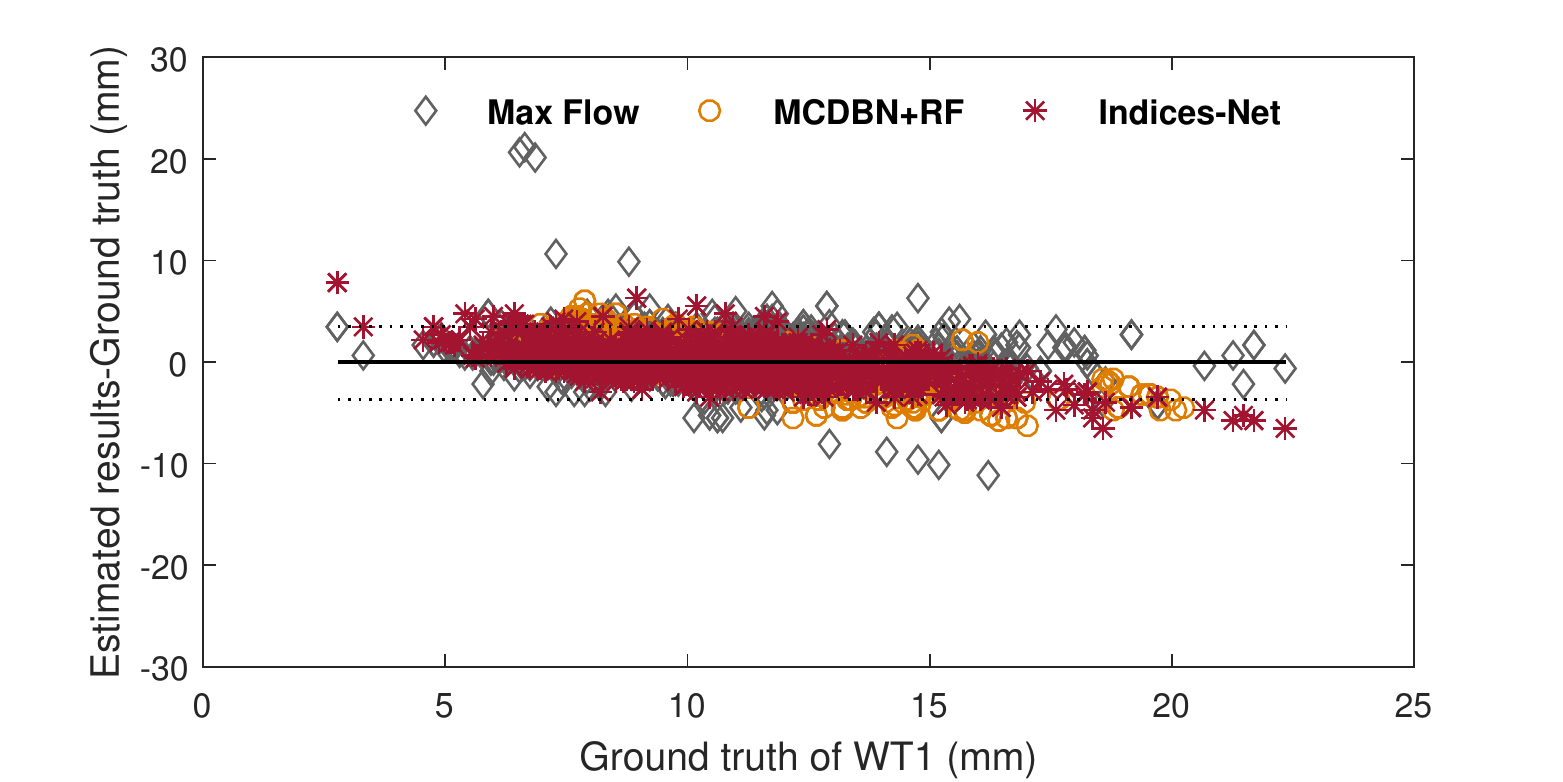}\label{fig_baplot_wt1}}
	\hfill
	\subfloat{\includegraphics[width=0.5\textwidth]{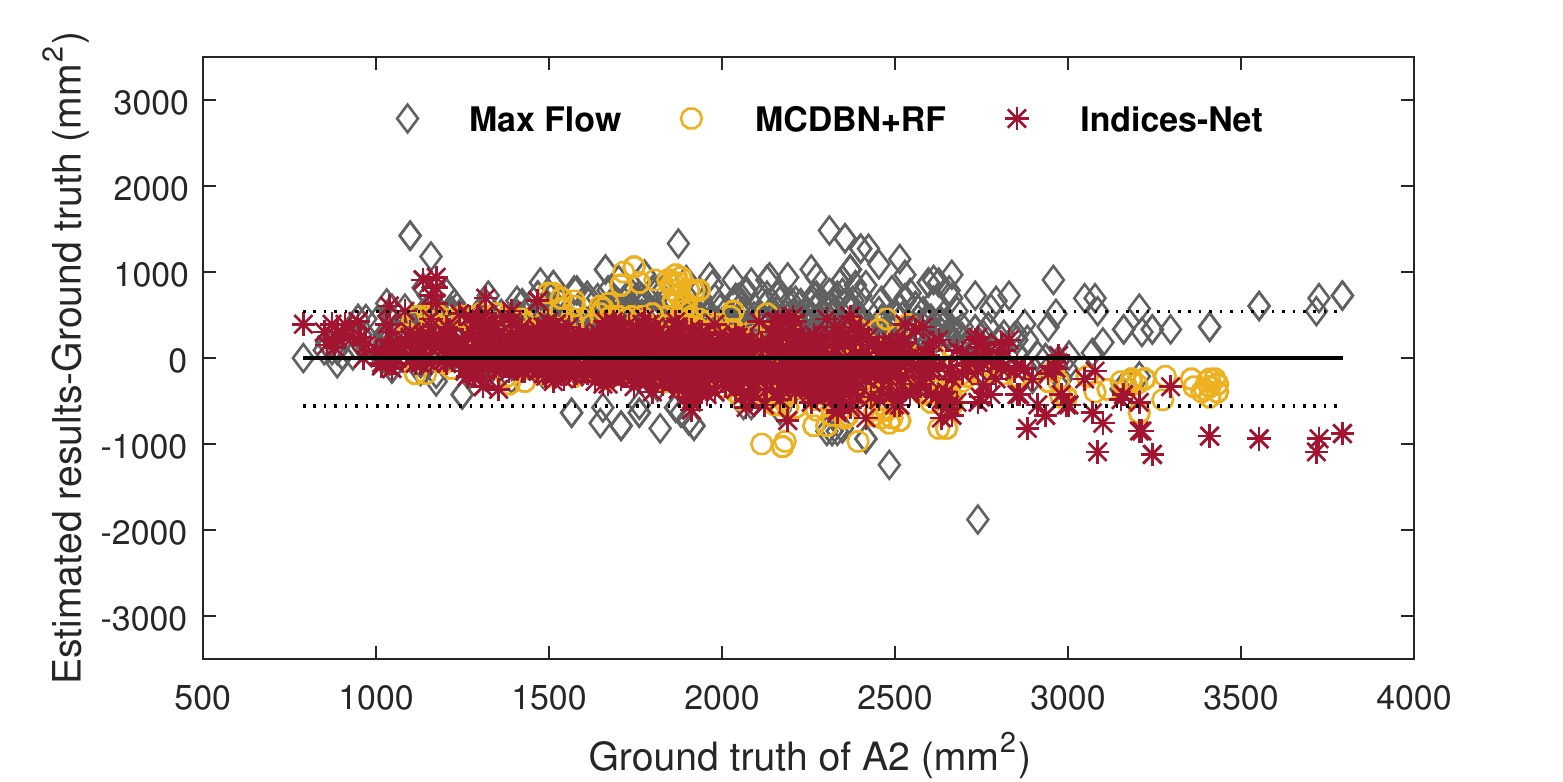}\label{fig_baplot_a2}}
	\caption{Bland-Altman plots for Max Flow, MCDBN+RF and Indices-Net. Only results of WT1 (left) and A2 (right) are illustrated for space reason. The horizontal dashed lines show the mean$\pm$1.96SD of the difference between estimations of Indices-Net and ground truth.}
	\label{fig_baplot}
\end{figure*}

\begin{figure*}[tbp]
	\centering
	\subfloat{\includegraphics[width=0.7\textwidth]{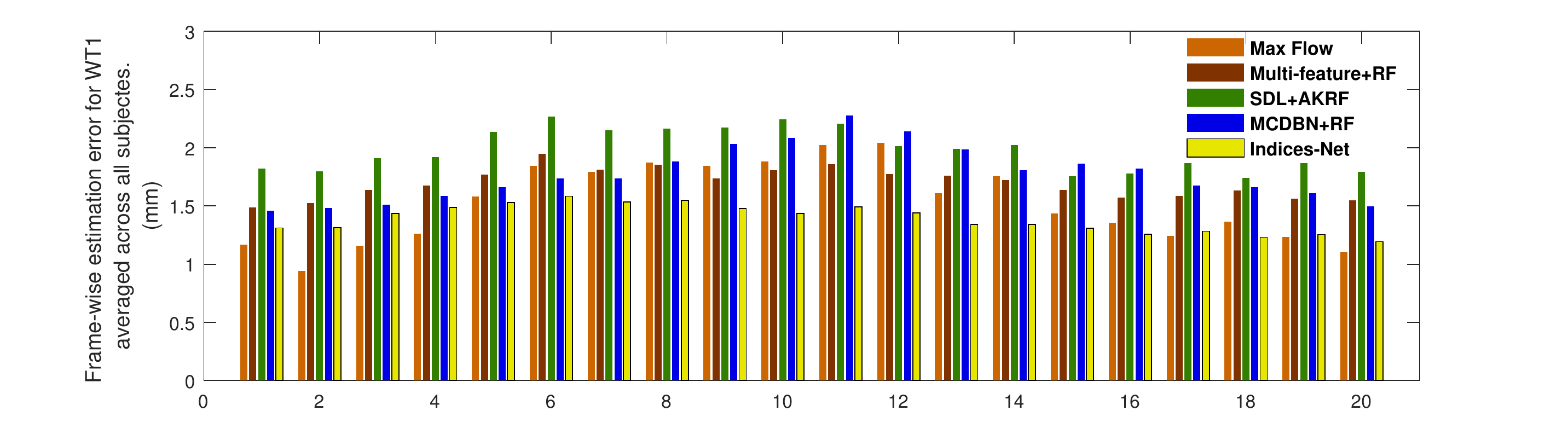}\label{fig_error_wt1}}
	\vfill
	\subfloat{\includegraphics[width=0.7\textwidth]{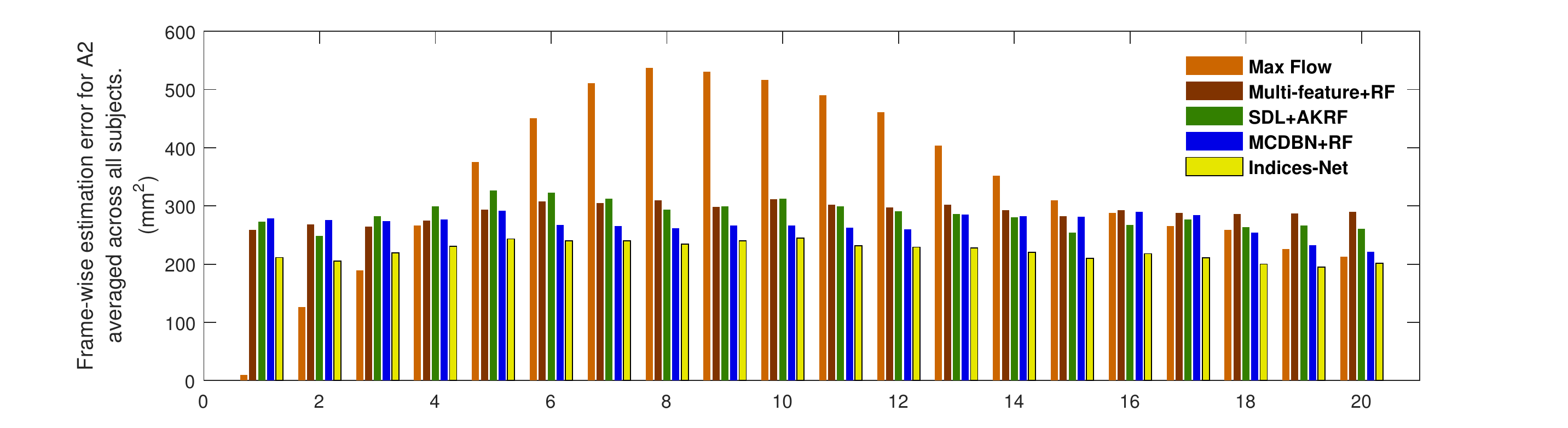}\label{fig_error_a2}}
	\caption{Frame-wise average absolute estimation error for WT1 (top) and A2 (bottom) over 145 subjects. Indices-Net achieves consistently lower estimation error for all frame than other competitors.}
	\label{fig_error}
\end{figure*}

\subsection{Experiments}
Extensive experiments are conducted to validate the effectiveness of our Indices-DCAE from the following aspects.

Firstly, performance of Indices-DCAE for multitype cardiac indices estimation and its effectiveness for cardiac function assessment are examined with our dataset following the five-fold cross validation protocol. 

Secondly, advantages of Indices-DCAE over existing segmentation-based and direct methods are extensively analyzed following the same five-fold cross validation protocol\footnote{The implementation of~\cite{ayed2012max} is from the original author, while the rest~\cite{zhen2014direct,zhen2015multi,zhen2015direct} are implemented by ourselves strictly following the original papers.}. Statistical significance of the better performance of Indices-DCAE is examined by one-tailed \emph{F}-test with significance level of 1\%. The test statistic is variance ratio $F=\frac{\sigma_0^2}{\sigma^2}$, where $\sigma_0^2$ and $\sigma^2$ are variances of the estimation errors (which is essentially the mean squared error given the fact that the mean value of estimation error is near zero) for Indices-Net and one competitor to compare with. A test result of $H=1$ indicates that Indices-Net achieves significantly lower estimation error variance than its competitor. 

Thirdly, benefit of joint learning is evidenced. Joint learning is capable of enhancing the expressiveness of the representation model and thus make it more compatible with the regression model. To demonstrate this, the expressiveness of two representations is compared in terms of cardiac indices estimation: 1) $F_{\text{Indices}}$, which is obtained from Indices-Net, as jointly learned feature; 2) and $F_{\text{Image}}$, which is obtained from Image-DCAE, as non-jointly learned feature. Both of them are computed from the 16 feature maps of the last deconvolution layer in DCAE by averaging values of all non-overlapping $5\times 5$ blocks in these feature maps, as the way in GIST descriptor~\cite{oliva2001modeling}, resulting a feature vector of length 4096 for each image. Once the two representation are available, random forest (RF) models with the same configuration (\emph{ntree}=1000, \emph{mtry}=200) are applied to them for cardiac indices estimation following the five-fold cross validation protocol.

Fourthly, benefit of our initialization strategy is evidenced by comparing the performance of Indices-DCAE with and without initialization from pre-trained DCAE.

Finally, two other deep networks for cardiac indices estimation are examined. 1) deep CNN, which contains only the convolution part of our DCAE followed by a additional fully connected layer for indices estimation. 2) FCN, which contains the first 10 convolution layers of FCN32s~\cite{FCN}, followed by convolution layers of $1024@5\times 5$, $1024@1\times 1$ and $8@1\times 1$ for indices estimation. Three models are examined for cardiac indices estimation: 1) deep CNN 1, which is trained from scratch; 2) deep CNN 2, which is finetuned from our pre-trained DCAE; 3) FCN, which is finetuned from PASCAL VOC-trained FCN-32s model (https://github.com/shelhamer/fcn.berkeleyvision.org).

\section{Results and Analysis}
In this section, we demonstrate the results of the above mentioned experiments to validate the effectiveness of Indices-Net in the task of multitype cardiac indices estimation, as well as its advantages over existing segmentation-based, two-phase direct methods, and other deep architectures.

\subsection{Estimation accuracy and effectiveness for cardiac function assessment}
The proposed Indices-Net achieves accurate estimation of all the cardiac indices, as shown in the last row of Table.~\ref{table_compare_wt}. Indices-Net estimates wall thicknesses with average MAE of 1.44$\pm$0.71mm, which is less than the mode of pixel spacing (1.5625mm/pixel) of the MR images in our dataset, and achieves average correlation of 0.758 with the ground truth. Among the six linear indices, the lateral wall thicknesses (WT3 and WT4) are more difficult to estimate due to the nearly invisible border between the lateral free wall of myocardium and the surroundings, and the presence of papillary muscle. Even so, Indices-Net is capable of obtaining accurate estimation with error of about one pixel, by leveraging the correlation between neighbouring indices with the index-wise connected CNN (see Fig.~\ref{fig_regscore5}). For the two planar indices, Indices-Net estimates both with low MAE (185$\pm$162mm$^2$ and 223$\pm$193mm$^2$) and high correlation (0.953 and 0.853). 

The effectiveness of Indices-Net for cardiac function assessment is also demonstrated in the last row of Table.~\ref{table_compare_function}. For all the 145 subjects, Indices-Net achieves estimation error of 6.22$\pm$5.01\% and correlation of 0.856 for ejection fraction, and (18.6$\pm$15.8\%, 0.610) for myocardium wall thickening. This the first time that automatic cardiac wall thickening estimation is studied.

\begin{table*}[!t]
	\caption{Left-tailed \emph{F}-test between Indices-Net and each of the competitor. Indices-Net significantly outperforms its competitors for nearly all the cases ($H=1$) at 1\% significance level. (CI: confidence interval.)}
	\label{table_Ftest}
	\centering
	\begin{tabular}{c||c|cccc||c|cccc}
		\hline
		Method&Index&\textit{H}&\textit{p}-value&CI& $\frac{\sigma_0^2}{\sigma^2}$&Index&\textit{H}&\textit{p}-value&CI& $\frac{\sigma_0^2}{\sigma^2}$\\
		\hline
		\multirow{4}{*}{Max Flow [3]}&WT1&1&$<0.01$&[0, 0.727]&0.667&WT5&1&$<0.01$&[0, 0.248]&0.227\\
		&WT2&1&$<0.01$&[0, 0.419]&0.384 &WT6&1&$<0.01$&[0, 0.509]&0.467\\
		&WT3&1&$<0.01$&[0, 0.424]&0.389 &A1&0&$0.323$&[0, 1.072]&0.984\\
		&WT4&1&$<0.01$&[0, 0.237]&0.217 &A2&1&$<0.01$&[0, 0.746]&0.684\\
		\hline
		\multirow{4}{*}{Multi-features+RF [9]} &WT1&1&$<0.01$&[0, 0.696]&0.638&WT5&1&$<0.01$&[0, 0.771]&0.707\\
		&WT2&1&$<0.01$&[0, 0.860]&0.788 &WT6&1&$<0.01$&[0, 0.598]&0.549\\
		&WT3&1&$<0.01$&[0, 0.804]&0.731 &A1&1&$<0.01$&[0, 0.714]&0.655\\
		&WT4&1&$<0.01$&[0, 0.805]&0.738 &A2&1&$<0.01$&[0, 0.634]&0.582\\
		\hline
		\multirow{4}{*}{SDL+AKRF [11]} &WT1&1&$<0.01$&[0, 0.542]&0.497&WT5&1&$<0.01$&[0, 0.666]&0.611\\
		&WT2&1&$<0.01$&[0, 0.835]&0.765 &WT6&1&$<0.01$&[0, 0.432]& 0.396\\
		&WT3&1&$<0.01$&[0, 0.810]&0.741 &A1&1&$<0.01$&[0, 0.958]&0.879\\
		&WT4&1&$<0.01$&[0, 0.724]&0.664 &A2&1&$<0.01$&[0, 0.646]&0.592\\
		\hline
		\multirow{4}{*}{MCDBN+RF [10]} &WT1&1&$<0.01$&[0, 0.729]&0.630&WT5&0&$0.117$&[0, 1.057]&0.944\\
		&WT2&1&$<0.01$&[0, 0.929]&0.802 &WT6&1&$<0.01$&[0, 0.738]&0.637\\
		&WT3&1&$<0.01$&[0, 0.916]&0.791 &A1&1&$<0.01$&[0, 0.882]&0.788\\
		&WT4&1&$<0.01$&[0, 0.933]&0.833 &A2&1&$<0.01$&[0, 0.804]&0.695\\
		\hline		
	\end{tabular}
\end{table*}

\subsection{Performance comparison}
Indices-Net reveals great advantages for cardiac indices estimation and cardiac function assessment, when being compared with segmentation-based and existing two-phase direct methods (Tables.~\ref{table_compare_wt} and~\ref{table_compare_function}).  

The average MAE reductions of Indices-Net over Max Flow~\cite{ayed2012max} are 55.1\% for the linear indices and 17.4\% for the planar indices, even though Max Flow achieved high dice metric (0.913) for LV cavity segmentation. When the epicardium border is involved, Max Flow fails to deliver accurate estimation, as shown by the results of wall thicknesses and myocardium area. Our further analysis reveals that the dependency on manual segmentation of the first frame makes the estimation error of Max Flow increase as the estimated frame becomes far from the first frame within the cardiac cycle (Fig.~\ref{fig_error}). This makes it incapable of LV function assessment such as wall thickening analysis. Indices-Net outperforms the best of existing direct methods with clear MAE reductions (12.7\%, 14.6\%) and correlation improvements (0.123, 0.079) for the linear and planar indices. This evidences that the two-step framework in these methods is not adequate to achieve accurate estimation for multitype cardiac indices. Figs.~\ref{fig_baplot} and~\ref{fig_error} reveal with more details that Indices-Net can deliver more robust and accurate estimation than its competitors. The Bland-Altman plots show that Max Flow is prone to overestimate cardiac indices, while MCDBN+RF overestimates small indices and underestimates large indices. On the contrary, Indices-Net is capable of estimating them with consistently low error. The bar plots of frame-wise estimation error also demonstrate that Indices-Net estimates cardiac indices with consistently low error for all frames of one cardiac cycle. When applied to cardiac function assessment, Indices-Net performs best considering both the estimation error and correlation for LV ejection fraction and wall thickening (Table.~\ref{table_compare_function}).

Table~\ref{table_Ftest} demonstrates the results of left-tailed \emph{F}-test for Indices-Net and other competitors for all the 8 indices. The test result $H$, \emph{p}-value, variance ratio and its confidence interval are demonstrated. The variance ratio and its confidence level show to which extent the proposed method differs from the competitors. Except the two cases where variance of estimation error of Max Flow for A1 and that of MCDBN+RF for WT5 are very close to those of Indices-Net, for all the rest cases, Indices-Net significantly outperforms these competitors.

\subsection{Benefit of joint learning framework}
From the results (Rows $F_{\text{Indices}}$+RF and $F_{\text{Image}}$+RF) shown in Table.~\ref{table_wt}, it can be drawn that $F_{\text{Indices}}$ achieves average MAE of 1.46mm and 207mm$^2$ for the linear and planar indices, versus 1.82mm and 272mm$^2$ obtained by $F_{\text{Image}}$. This evidences that joint learning makes $F_{\text{Indices}}$ more expressive with respect to cardiac indices, therefore leads to lower estimation error and better correlation.

\begin{table}[t]
	\caption{Effects of pre-training DCAE and joint learning for cardiac indices estimation. Average performance is demonstrated here for the linear and planar indices.}
	\label{table_wt}
	\centering
	\setlength{\tabcolsep}{4pt}
	\begin{tabular}{*c||c|c|c|c}
		\hline
		\multirow{2}{*}{Method}&\multicolumn{2}{c|}{linear indices (mm)}&\multicolumn{2}{c}{planar indices (mm$^2$)}\\
		\cline{2-5}
		&MAE&$\rho$&MAE&$\rho$\\
		\hline
		$F_{\text{Image}}$+RF&1.82$\pm$1.05&0.577&272$\pm$168&0.819\\
		\hline
		$F_{\text{Indices}}$+RF&1.46$\pm$0.70&0.751&207$\pm$143&0.897\\
		\hline
		Indices-Net (scratch)&1.68$\pm$0.91&0.649&258$\pm$152&0.845\\
		\hline
		Indices-Net &1.44$\pm$0.71&0.758&204$\pm$133&0.903\\
		\hline
	\end{tabular}
\end{table}

\begin{figure}[t]
	\centering
	\includegraphics[width=9cm]{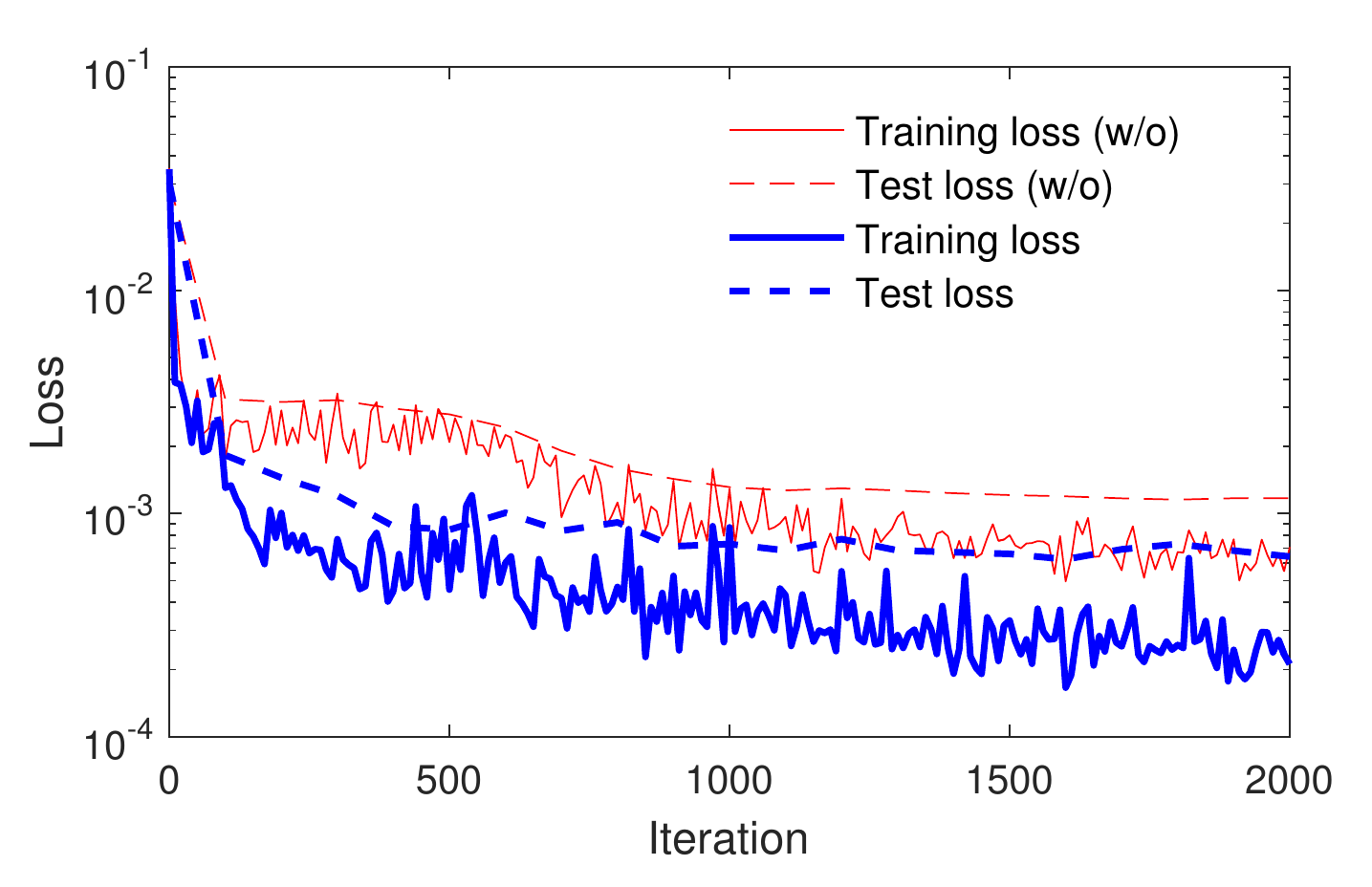}
	\caption{Curves of estimation loss during the training and test procedures with (blue) and without (red) pre-training, vs. the number of iteration. Pre-trained DCAE helps the network train faster and achieve lower estimation error for both training and test procedures.}
	\label{fig_loss_curve}
\end{figure}

\subsection{Benefit of initialization with pre-trained DCAE} \label{seg_benefit_pretrain}
The loss curves (Fig.~\ref{fig_loss_curve}) of the proposed Indices-Net with and without pre-training of the DCAE network clearly show that pre-trained DCAE helps the network train faster and converge to lower estimation error for the training procedure, and generalize better to the test dataset. Comparing the performance of Indices-Net and Indices-Net (scratch) in Table.~\ref{table_wt}, we can draw that without pre-training, the deep network fails to deliver accurate prediction for both types of cardiac indices. With the pre-trained DCAE as initialization, our Indices-Net is capable of disentangling index-relevant features and delivering accurate estimation for multitype cardiac indices.

\subsection{Performance comparison with other deep networks}
Table~\ref{table_deep_network} demonstrates that our Indices-Net achieves much lower MAE than the single deep CNN and the FCN networks in estimation of the two types of cardiac indices. The superior performance of Indices-Net is mainly contributed by the DCAE for representation and the index-wise connection of CNN for regression. Both the discriminative encoder and the generative decoder in DCAE builds a cascade mapping of \emph{cardiac image (input layer)$\rightarrow$ latent representation (fc6)$\rightarrow$ indices-relevant structures}. Then the index-wise connection of CNN effectively disentangles the most relevant index-specific features from these structures, while keeping inter-indices correlation.

\begin{table}[t]
	\caption{Performance comparison of Indices-Net with other deep networks for cardiac indices estimation. Average MAE performance is demonstrated here for the linear and planar indices.}
	\label{table_deep_network}
	\centering
	\setlength{\tabcolsep}{4pt}
	\begin{tabular}{*c||c|c}
		\hline
		Method&linear indices (mm)&planar indices (mm$^2$)\\
		\hline
		deep CNN 1&1.96$\pm$1.00&301$\pm$180\\
		\hline
		deep CNN 2&1.94$\pm$0.95&297$\pm$180\\
		\hline
		FCN &2.07$\pm$0.94&339$\pm$157\\
		\hline
		Indices-Net&1.44$\pm$0.71&204$\pm$133\\
		\hline
	\end{tabular}
\end{table}

\section{Conclusion and Discussion}
A deep integrated network Indices-Net was proposed to estimate frame-wise multitype cardiac indices simultaneously and achieved highly reliable and accurate estimation for all the cardiac indices when validated on a dataset of 145 subjects. Jointly learning of the two tightly-coupled networks DCAE and CNN enhanced the expressiveness of image representation and the compatibility between the indices regression and image representation. It is the first time that multitype cardiac indices estimation is investigated and the first time that joint learning of image representation and indices regression is deployed in cardiac indices estimation. 

The success of the proposed method for mid-cavity slice paved a great way to the true 3D (multi-slice) estimation of cardiac indices, which is usually used in clinical application. To achieve 3D estimation of multitype cardiac indices, Indices-Net can be directly trained with multi-slice cardiac MR images as in existing CNN-based 3D volume estimation~\cite{kabani2016estimating}, or adapted with the recurrent neural network to model the dependencies of neighbouring slices in the latent space.


\ifCLASSOPTIONcaptionsoff
  \newpage
\fi

\bibliographystyle{IEEEtran}
\bibliography{lv_wall}


%







\end{document}